\documentclass{article}

\usepackage{microtype}
\usepackage{graphicx}
\usepackage{subfigure}
\usepackage{booktabs} 
\usepackage{graphicx}

\usepackage[utf8]{inputenc} 
\usepackage[T1]{fontenc}    
\usepackage{hyperref}
\usepackage{url}            
\usepackage{booktabs}       
\usepackage{amsfonts}       
\usepackage{nicefrac}       
\usepackage{microtype}      
\usepackage{color}
\usepackage{amsmath}
\DeclareMathOperator*{\argmax}{arg\,max}

\usepackage{enumitem} 
\usepackage{bm}

\usepackage{hyperref}

\usepackage{caption}

\usepackage[accepted]{icml2019}

\icmltitlerunning{Dynamic Measurement Scheduling for Event Forecasting Using Deep RL}

\begin{document}

\twocolumn[
\icmltitle{Dynamic Measurement Scheduling for Event Forecasting Using Deep RL}



\icmlsetsymbol{equal}{*}

\begin{icmlauthorlist}
\icmlauthor{Chun-Hao Chang}{equal,to,vec,sk}
\icmlauthor{Mingjie Mai}{equal,to,vec,sk}
\icmlauthor{Anna Goldenberg}{to,vec,sk}
\end{icmlauthorlist}

\icmlaffiliation{to}{University of Toronto, Toronto, ON, Canada}
\icmlaffiliation{vec}{Vector Institute, Toronto, ON, Canada}
\icmlaffiliation{sk}{The Hospital for Sick Children, Toronto, ON, Canada}

\icmlcorrespondingauthor{Chun-Hao Chang}{kingsley@cs.toronto.edu}

\icmlkeywords{Machine Learning, ICML}

\vskip 0.3in
]



\printAffiliationsAndNotice{\icmlEqualContribution} 

\begin{abstract}

Imagine a patient in critical condition.
What and when should be measured to forecast detrimental events, especially under the budget constraints?
We answer this question by deep reinforcement learning (RL) that jointly minimizes the measurement cost and maximizes predictive gain, by scheduling strategically-timed measurements. 
We learn our policy to be dynamically dependent on the patient's health history. 
To scale our framework to exponentially large action space, we distribute our reward in a sequential setting that makes the learning easier.
In our simulation, our policy outperforms heuristic-based scheduling with higher predictive gain and lower cost. 
In a real-world ICU mortality prediction task (MIMIC3), our policies reduce the total number of measurements by $31\%$ or improve predictive gain by a factor of $3$ as compared to physicians, under the off-policy policy evaluation.

\end{abstract}

\section{Introduction}

Redundant and expensive screening procedures and lab measurements have increased the overall health care costs \citep{feldman2009managing}. 
This phenomenon, either due to commercial interests or over-concern, has been observed in numerous clinical practices \citep{hoffman2012overdiagnosis, brodersen2018overdiagnosis}. 
For example, numerous studies \citep{iosfina2013implementation, pageler2013embedding} found no evidence that regular blood testing improves diagnosis; frequent blood tests may even cause anemia and infection \citep{salisbury2011diagnostic}.
To combat the situation, \citet{dewan2017reducing} devised a simple rule to reduce the frequency of blood tests by $87\%$ in pediatric ICU. 
Similarly, \citet{kotecha2017reducing} showed that the measurement costs can be significantly reduced without increase in mortality or re-admission rates in cardiac and surgical ICU.
These findings point toward the need for principled data-driven approaches for lab test scheduling to improve the healthcare system.

Recently developed time-series forecasting models solve the much needed problem of early detection of adverse events (e.g. sepsis) based on sparse and irregular measurements \citep{ghassemi2015multivariate, soleimani_scalable_2017,futoma_improved_2017}. However, the timing of these measurements varies from doctor to doctor and from one hospital to another, leading to a drastically different input distribution that may result in inferior classifier performance. Additionally, these classifiers are not often built to provide insights into which measurements help the most to make the prediction given current patient's condition. 

We propose a scalable and flexible framework that learns a data-driven and dynamic sampling policy using deep Q-learning. Deep Q-learning, a type of Reinforcement  Learning (RL), is a powerful framework that can learn from large amount of retrospective data even when the data does not represent optimal behaviors. In addition, it has been shown to be promising for solving various clinical problems \citep{raghu2017continuous, futoma2018learning}.

Our framework is a two-tier system. 
First, we learn an event forecasting model to represent the patient's condition.
Then we train RL to maximize this model's performance while minimizing the cost of the needed measurements. 
Compared to directly using the event as reward, our approach of using event probabilities from a learned classifier gives the RL immediate reward for every action, making reward assignment and training comparably easier. 
To handle the exponentially large action space in the measurement scheduling problems, we use sequential action setting that successfully handle the number of measurements in an unprecedented scale.

We show that in the simulation setting, when given a near-perfect classifier, our method is able to learn a strategically timed measurement scheduling that outperforms all the heuristic-based scheduling.
We then test it on MIMIC3, a real ICU temporal dataset. 
We compare our learned policies, physician's policy, as well as random policies using off-policy policy evaluation (OPPE) method, showing that our learned policies reduce measurement costs by $31\%$ or increase information gain by a factor of $3$ compared to physician's policy. 
Our data preprocessing and code are available online at  \url{https://github.com/zzzace2000/autodiagnosis}. 



\begin{figure*}
  \includegraphics[width=1.0\textwidth]{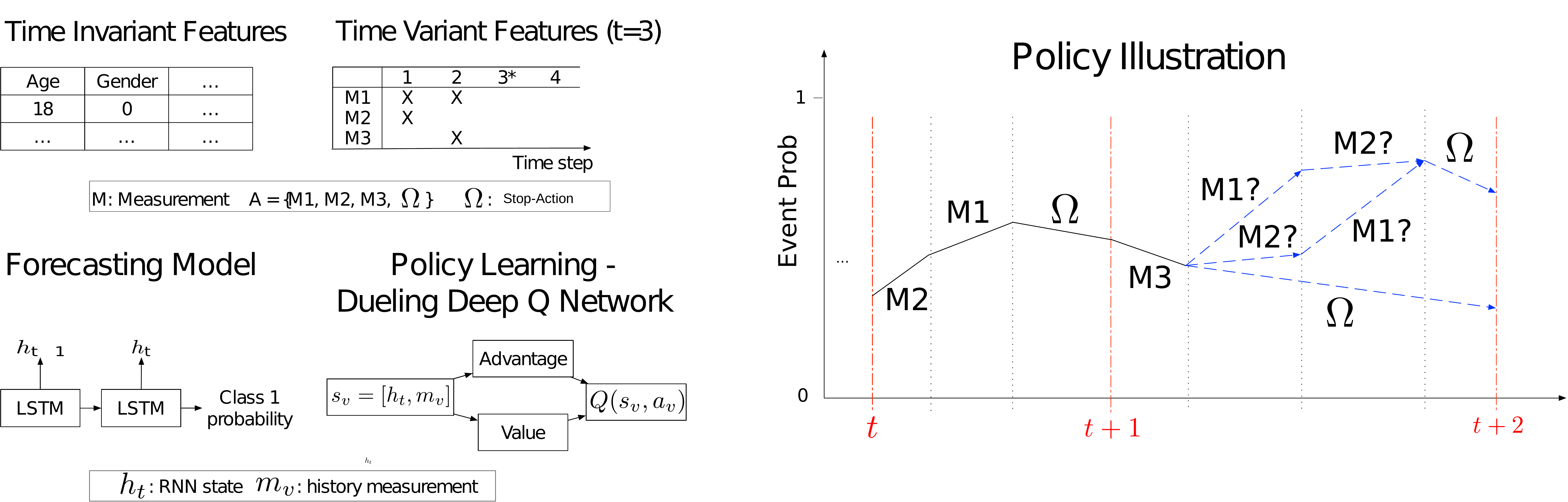}
  \caption{Our System Pipeline. (\textbf{Left}) Given a medical dataset with time-invariant and variant features, we train a forecasting model LSTM which produces an event (e.g. mortality) probability. 
  Then we train a dueling deep Q network to maximize the event probability and minimize number of measurements. Its input $s_v$ is the concatenation of LSTM hidden state $h_t$ (summarizing the past information) and the one-hot encoded measurements $m_v$ already made at this time.
  (\textbf{Right}) Policy illustration. The agent sequentially decides whether to take another measurement (M1, M2, M3) or stops making more measurements ($\Omega$) at the current timepoint.}
  \label{fig:overall_pipeline}
\end{figure*}

\section{Related Work} 



\subsection{Clinical Event Forecasting Models} Several models have been proposed for event forecasting on irregularly sampled EHR data. 
\citet{zhang2017medical} first used a deep generative variational recurrent neural network (VRNN) to learn feature representation and then used a neural network to predict disease.
\citet{li2016scalable, futoma_improved_2017} used multi-output Gaussian process (MGP) to impute the irregularly-sampled time series data on the grid points and used those to make predictions via recurrent neural network (RNN). 
\citet{soleimani2018scalable} also used a MGP to impute the missing data, but instead uses a survival model to predict the disease.


\subsection{Deep RL in healthcare} 
Several recent works use RL to learn a treatment plan in ICU.
\citet{weng2017representation} uses Q-learning to address glycemic control problem for sepsis patients.
\citet{prasad2017reinforcement} also uses Q-learning to recommend personalized sedation dosage and ventilator support.
\citet{raghu2017continuous} and \citet{komorowski2018artificial} focuses on treatments for sepsis using Q-learning. The action space is discretized over doses of two drugs commonly given to septic patients. 
\citet{futoma2018learning} improves the Q-learning model by MGP to impute the missing value and adopts RNN as the Q-learning network. 
\citet{wang2018supervised} learns a safe treatment scheduling policy that both matches existing physician policy and maximizes long-term reward using actor-critic framework. However, this approach is less meaningful when physician policy is sub-optimal, which may be the case for measurement scheduling. 
All the RL frameworks in healthcare above focus on the treatment scheduling problem.
Moreover, they either consider the effect of action to be independent or allow only a few actions to be scheduled. 
Our framework is scalable to large number of actions and consider multiple actions jointly, which is a more realistic setting in the clinical practice. 


Beside treatment scheduling, \citet{Cheng2018} also aims to learn a measurement policy by RL. 
They use fitted Q-iteration to schedule $4$ different lab tests relevant to diagnosis of sepsis and acute renal failure in the ICU setting. 
Our work differs in three main ways. 
First, they treat the scheduling of each measurement independent, making it unsuitable in ICU since the lab measurement values are highly correlated and sampling policy should be considered jointly across all measurements.
We show that this independent design underperforms substantially than our sequential design policy in section \ref{sec:mimic}, and could be the reason for their sometimes subpar performance against random policy. 
Second, their reward is different from ours: they design a multi-objective reward such as SOFA score or missingness, while we represent the informativeness of a new measurement using a trained classifier as our reward and use linear combination to combine multiple objectives.
Third, their MDP state formulation doesn't explicitly capture historical information of the patient. Instead, our work summarizes the historical information trends explicitly using LSTMs, and shares this representation both for risk scoring and action choosing.



\subsection{Active feature acquisition}
Several works \citep{contardo2016sequential, he2016active, shim2017pay} study the problem of selecting a subset of features to achieve the maximum prediction performance for a non-time-series classifier. We tackle time-series feature acquisition problem where historical information matters. This is especially true in a healthcare setting. In addition, being time-series, the choice of a measurement at the current timepoint affects the performance of the prediction model at a future timepoint. 

\subsection{Active sensing in medical setting}
The focus of active sensing is to determine what and when to measure when acquiring measurements is costly. 
\citet{ahuja_dpscreen_2017} handles single-measurement scheduling problem for breast cancer screening by adopting a fixed model-based transition model. Unfortunately, it requires strong assumption, knowing the disease model dynamics, and does not handle multiple types of measurements. 
Similarly, \citet{yoon2018deep} proposes a method of scheduling measurements to trade between uncertainty in prediction and the measurement cost. 
Their model performs a measurement for the next time stamp if the decreases in the uncertainty in prediction exceed the measurement cost. 
Our approach differs in three ways. 
First, we use Q-learning to learn policy that maximizes cumulative discounted reward of patient trajectories, while they greedily select measurements that would exceed the utility threshold at the next time stamp. 
Second, we consider a different definition of informativeness of a new measurement - gain in predictive probability. Consider a binary case, where the model produces a wrong estimate, a measurement that encourages a lower uncertainty would not be the ideal choice of action. 
Third, at test time, instead of evaluating reward at run time, our RL agent speeds up the computation by amortized inferring the corresponding Q-value by the learned Q function. 

\section{Methods}


Our framework is composed of two parts: a forecasting predictive model and a RL model.
See Figure \ref{fig:overall_pipeline} for an overview.
For the first part, we train a multi-layer LSTM classifier testing \citep{hochreiter1997long} to forecast events of interest using various features.
We then frame measurement scheduling question as a sequential feature acquisition problem by RL. 
We train a dueling deep Q-learning network (DQN) to schedule measurements that maximizes the classifier's predictive probability while lowering measurement cost given patient's history up to the given timepoint.

\subsection{Deep LSTM Classifier}
To handle the sparse time-series data in LSTM, we use mean imputation to fill in the missing measurement values.
We concatenate the imputed measurement values with missingness indicators and the static demographics for each timepoint $t$ and individual $i$. 
To learn the classifier $\mathcal{I}$, we minimize cross entropy loss between RNN's prediction and true label by backpropogation (Figure \ref{fig:overall_pipeline}, Forecasting Model). 
We list all the hyperparameters in appendix \ref{appendix:classifier_performance}.

\subsection{Dueling Deep Q Network (DQN)}
Dueling DQN factorizes the computation of Q-value into value stream and advantage stream \citep{wang2015dueling}, i.e.
\begin{equation}
    Q(s, a) = V_{\eta}(f_{\xi}(s)) + A_{\psi}(f_{\xi}(s), a) - \frac{\sum_{a'}A_{\psi}(f_{\xi}(s), a')}{N_{action}}
\end{equation}
where $\xi$, $\eta$, and $\psi$ are respectively, the parameters of the shared encoder $f_{\xi}$ of the value stream $V_{\eta}$, and of the advantage stream $A_{\psi}$ (Figure \ref{fig:overall_pipeline}, Policy Learning).

\paragraph{Sequential Actions Design}
In our clinical data, lots of measurements have the exact same time for convenience, i.e. there is no known true scheduling order. 
Given $K$ possible measurements, at any given time the agent has to decide among $2^K$ large combinations of measurements, which is clearly unscalable to large $K$. 
In addition, naively assigning reward to a set of actions without considering the commonality between sets of actions lead to more difficult learning and gets lower sample efficiency.
To overcome these two difficulties, we design the RL to take actions in a sequential manner to overcome the large action space and assign separate reward to each individual action (Figure \ref{fig:overall_pipeline}, Right).
Specifically, we include a new action $\Omega$ to represent stopping making any more action.
Then at each time point, the agent chooses the action with maximum Q-value one at a time until the agent selects action $\Omega$ (Algorithm \ref{alg:execution}). 

\begin{algorithm}[tb]
   \caption{Running policy}
   \label{alg:execution}
\begin{algorithmic}
   \STATE {\bfseries Input:} LSTM hidden state $h_t$, policy $Q$ \\
   \textbf{Output:} DQN actions $A_t$ \\
   Initialize actions $A_t = \emptyset$ \\
   \WHILE{$\Omega \not\in$ $A_t$}
   \STATE 
      $s_t \leftarrow [h_t, A_t]$ \\
      $a \leftarrow \argmax_{a' \not\in A_t} Q(s_t, a')$ \\
      Add $a$ into $A_t$ \\
   \ENDWHILE \\
\end{algorithmic}
\end{algorithm}

\textbf{Action}
We add a new stop-action $\Omega$ into RL actions. 
We represent RL agent's action $a_v$ as a multi-hot encoding vector of size $K + 1$. For $k \in [1,K]$, $a_{v, k}=1$ denotes the $k^{th}$ measurement is taken at this timepoint, otherwise $a_{v, k}=0$. 

\textbf{Reward}
We define the reward function as a linear combination of the information gain $g_{\mathcal{I}}$ and measurement cost $c$, i.e. $r(s_v, a_v) = g_{\mathcal{I}}(s_v, a_v) - \lambda * c(a_v)$, where $v$ represents the step in the MDP (to differentiate between timepoint $t$).
To encourage the predictive performance of the classifier $\mathcal{I}$, we define the information gain $g(s_v, a_v)$ as the probability change of the classifier $\mathcal{I}$, conditioned on the label, i.e.

\begin{equation}
    g_{\mathcal{I}}(\Delta_P) =
    \begin{cases}
      \Delta_P , & \text{if}\ label=1 \\
      -\Delta_P, & \text{otherwise}
    \end{cases}
    \label{eq:info_gain}
\end{equation}

The cost of scheduling a measurement $c(a_v)$ is a hyperparameter and should be defined by the domain expert which could represent its monetary cost, operational complexity or patient's discomfort. 
In this work we simply define it as the number of measurements except the action $\Omega$ i.e.
\begin{equation}
    c(a_v) =
    \begin{cases}
      1, & \text{if}\ a_v \neq \Omega \\
      0, & \text{otherwise}
    \end{cases}
\end{equation}

\textbf{State}
We use a multi-hot encoding $m_v$ to denote the measurements that have been scheduled by the agent at the current timepoint.
We use the concatenation of last LSTM layer representation $h_t$ of patient's history and history measurement $m_v$ as the input to the agent, denoted $s_v = [h_t, m_v]$.


\begin{algorithm}[tb]
   \caption{Generate experience for a patient at time $t$}
   \label{alg:exp_generation}
\begin{algorithmic}
    \STATE {\bfseries Input:} 
    Pretrained LSTM model $\mathcal{I}$, 
    current observation $y_t = \{y_{t, 1},...,y_{t, K_t}\}$, patient's history observations $q_{t-1} = \{y_1,..., y_{t-1}\}$, 
    decay factor $\gamma$, 
    total number of measurement $K$,
    action cost scale factor $\lambda$. 
    \\ $h_{\mathcal{I}}(q, x), p_{\mathcal{I}}(q, x)$: last hidden state and the probability of $\mathcal{I}$ with patient's observations $q$ and prediction time $x$ \\
    \textbf{Output:} All training experiences tuple $E$ \\

    \vspace{3pt}
    $E = \emptyset$ \\ 
    Store time-passing experience from from $t-1$ to $t$ \\ 
    
    [$h = h_{\mathcal{I}}(q_{t-1}, t-1)$,
    $m = y_{t-1}$, 
    $h' = h_{\mathcal{I}}(q_{t-1}, t)$, 
    $m'=\emptyset$, $a = \Omega$, 
    $r = (p_{\mathcal{I}}(q_{t-1}, t) -p_{\mathcal{I}}(q_{t-1}, t-1))$, 
    $\gamma=\gamma$] in $E$ \\
    \vspace{3pt}
    Randomly shuffle $y_t$
    \FOR{$v=1$ \textbf{to} K}
    \STATE
      Store measurement experience [$h = h_{\mathcal{I}}(q_{t-1}, t)$, $m=\{y_{t,1},...,y_{t, v-1}\}$, 
      $h' = h_{\mathcal{I}}(q_{t-1}, t)$, $m'=\{y_{t, 1}...y_{t, v}\}]$, 
      $a = index(y_{t,v})$, 
      $r = (p_{\mathcal{I}}(q_{t-1} \cup \{y_{t,1},...,y_{t, v-1}\}, t) - p_{\mathcal{I}}(q_{t-1} \cup \{y_{t, 1},...,y_{t,v}\}, t) - \lambda * c(a))$, 
      $\gamma=1$] in $E$
    \ENDFOR
   
\end{algorithmic}
\end{algorithm}

\textbf{Learning}
We generate RL experience tuples [$h$, $m$, $h'$, $m'$, $a$, $r$, $\gamma$] in a sequential manner (Algorithm \ref{alg:exp_generation}).
We generate two kinds of experience, time-passing experiences and measurement experiences. 
The time-passing experience assigns the probability change due to time shift from $t-1$ to $t$ to the action $\Omega$.
The measurement experience assigns the reward to a specific measurement action.
Since multiple measurements are recorded at the same time and we do not know the underlying chronological order, we thus treat every order equally likely.
We generate training experience based on several random order of the measurements at the same time point $t$, as a way of data augmentation. 
For example, if $M1, M2, M3$ were recorded at a timepoint, the action order could be (M1, M2, M3, $\Omega$), (M2, M1, M3, $\Omega$), or (M3, M2, M1, $\Omega$) etc. 
To avoid the total reward received change across different random orders, we do not decay the reward ($\gamma=1$) in these experiences.
Under our linear additive reward design (eq. \ref{eq:info_gain}), the random order produces the same culmulative reward no matter which order is used. 
Also, we do not update the hidden state $h$ for these measurement experiences within the same time $t$ since we do not know the measurement until $t+1$.

We optimize the RL agent by minimizing the Bellman-equation square error (Algorithm \ref{alg:training}). 
Note that when calculating the $Q_{target}$, the best action considered can not be in the action set $m'$ already performed in the current time. i.e. 
$$ Q_{target}(s, a, s') = r(s, a) + \gamma \max_{a' \not\in m'} Q(s', a') $$
All the training hyperparameters are listed in Table \ref{table:dqn_hyperparameters}.

\begin{algorithm}[tb]
   \caption{Training sequential DQN}
   \label{alg:training}
\begin{algorithmic}
    \STATE {\bfseries Input:} Pretrained LSTM model $\mathcal{I}$, patient's database $D = \{q^1,...,q^N \}$, patient's trajectory length $T^i$. \\
   \textbf{Output:} DQN model $Q_{\theta}$

    
    $R \leftarrow \emptyset$ // Initialize prioritized experience replay buffer $R$
    \FOR{$q^i$ in $D$}
        \FOR{$t=1$ \textbf{to} $T^i$}
            \STATE 
            $E^i \leftarrow$ get experience for patient $q^i$ at $t$ (Algo. \ref{alg:exp_generation}) \\
            Store $E^i$ in $R$
        \ENDFOR
    \ENDFOR
    
    \WHILE{$L$ is not converged}
        \STATE  
        $E\sim R $ \\
        $h, m, h', m', a, r, \gamma \leftarrow E $\\
        $s = [h, m], s' = [h', m']$ \\
        $Q_{target}(s, a, s') = r(s, a) + \gamma \max_{a' \not\in m'} Q_{\theta}(s', a')$ \\
        $\textrm{minimize\ } L = [Q_{\theta}(s, a) - Q_{target}(s, a, s')]^2$ \\
        Update priority of $E$ in $R$ using $L$ 
    \ENDWHILE
   
\end{algorithmic}
\end{algorithm}

\begin{algorithm}[tb]
  \caption{Per-time off-policy evaluation}
  \label{alg:offpolicy_evaluation}
\begin{algorithmic}
    \STATE {\bfseries Input:} Trained value estimator regression model $\phi$, patient's database $D = \{q^1...q^N \}$, DQN state $s_{t}^{i}$, trained DQN agent $Q$. \\
  \textbf{Output:} Estimated cumulative information gain $G$

    $G = 0$
    \FOR{$q^i$ in $D$}
        \FOR{$t=1$ \textbf{to} $T^i$}
            \STATE 
        $a_t^Q \leftarrow$ run $Q$ with patient state $s_{t}^{i}$ (Algo. \ref{alg:execution}) \\
            $\Delta_p^Q = \phi(s_t, a_t^Q)$ // Estimate probability changes \\
            $G = G + \gamma^t * g_{\mathcal{I}}(\Delta_p^Q)$
        \ENDFOR
    \ENDFOR
\end{algorithmic}
\end{algorithm}

\subsection{Off-Policy Policy Evaluation (OPPE)}
OPPE is currently the only way to evaluate RL performance on retrospective data, and is crucial to report OPPE to make it possible for a future online evaluation in a healthcare setting. 
We use regression-based estimator \citep{jiang2015doubly} to estimate the values of physician and our learned policies using physician collected data.
We do not use importance-sampling based method since it would require an exact match with physician actions under our deterministic policy, which is virtually impossible in our high-dimensional action space.
Besides, it is also shown to be unstable when using with regression-based estimator in \citet{Liu2018RepresentationBM}.

We use per-time value estimator to evaluate our learned policies (Algorithm \ref{alg:offpolicy_evaluation}).
First, we train a regression model $\phi$ that maps the state-action pair to the information gain probability changes of model $\mathcal{I}$. 
Specifically, at each time $t$, the input is the concatenation of the latent state $h_t$ and multi-hot encoding of actions $a_t$ performed at time $t$, and the output is the probability changes $\Delta_P = P_{t+1} - P_t$.
We use feed-forward neural network to fit the regression with all hyperparameters listed in Appendix Table \ref{table:info_gain_est_hyperparameters}.
Then, for each patient at each time $t$, we estimate to the next time $t+1$ what is the corresponding reward if the specified action is performed. 
And we obtain estimated cumulative information gain $G$ by summing over all estimated information gain $g_{\phi}$ across all patients and all time $t$ with decay as $\gamma^t$.

\section{Results}


\begin{figure}[t]
  \centering
  \includegraphics[width=0.4\textwidth]{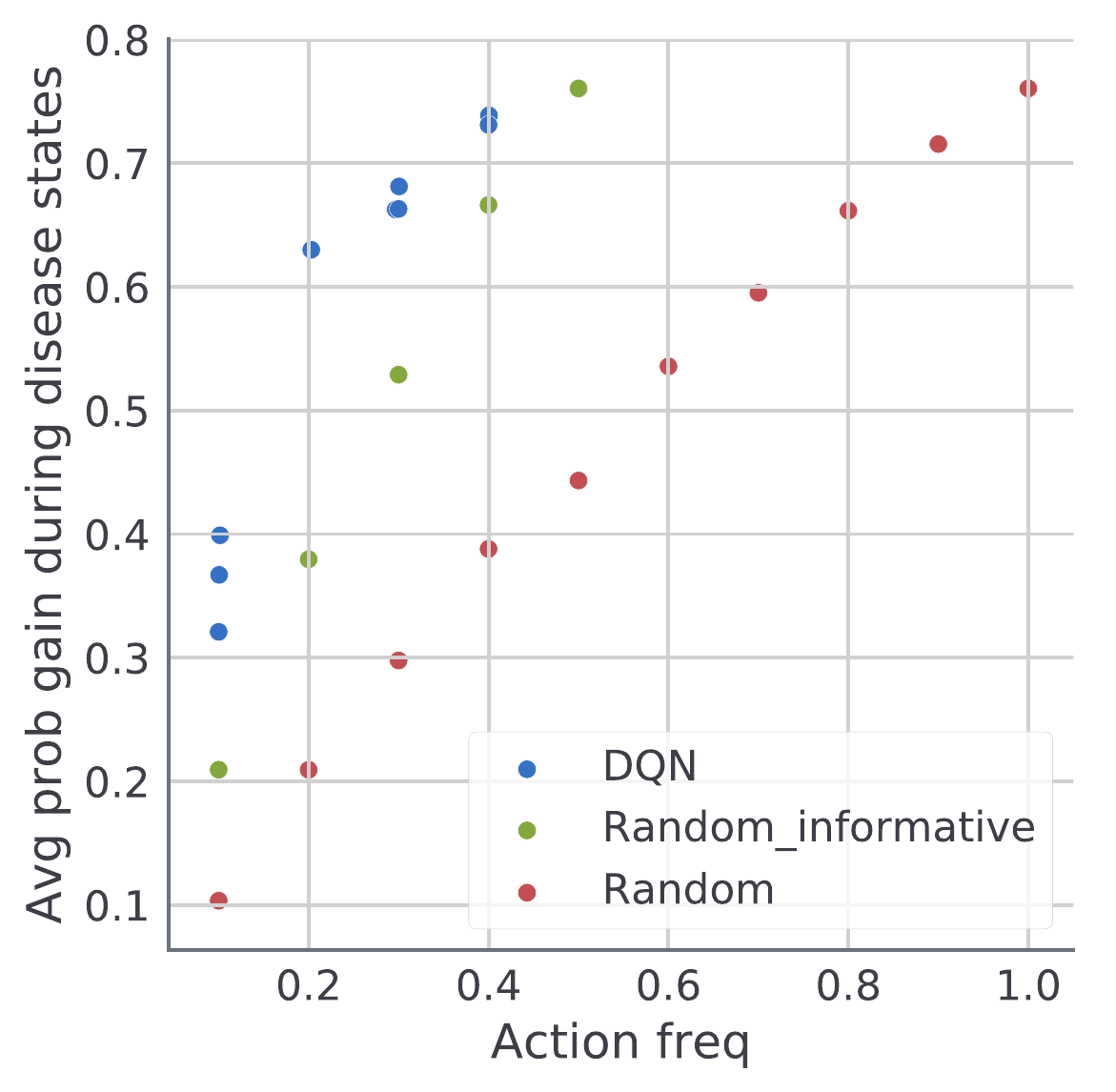}
  \caption{Online evaluation of policies in simulation. Action freq is the number of measurements taken average over all trajectories. An ideal policy should have low action frequency and high probability gain during disease state (i.e. top left corner).}
  \label{fig:reward_roc_plot_simulation}
\end{figure}

\subsection{Simulation}

\begin{figure}[t]
  \centering
  \includegraphics[width=0.5\textwidth]{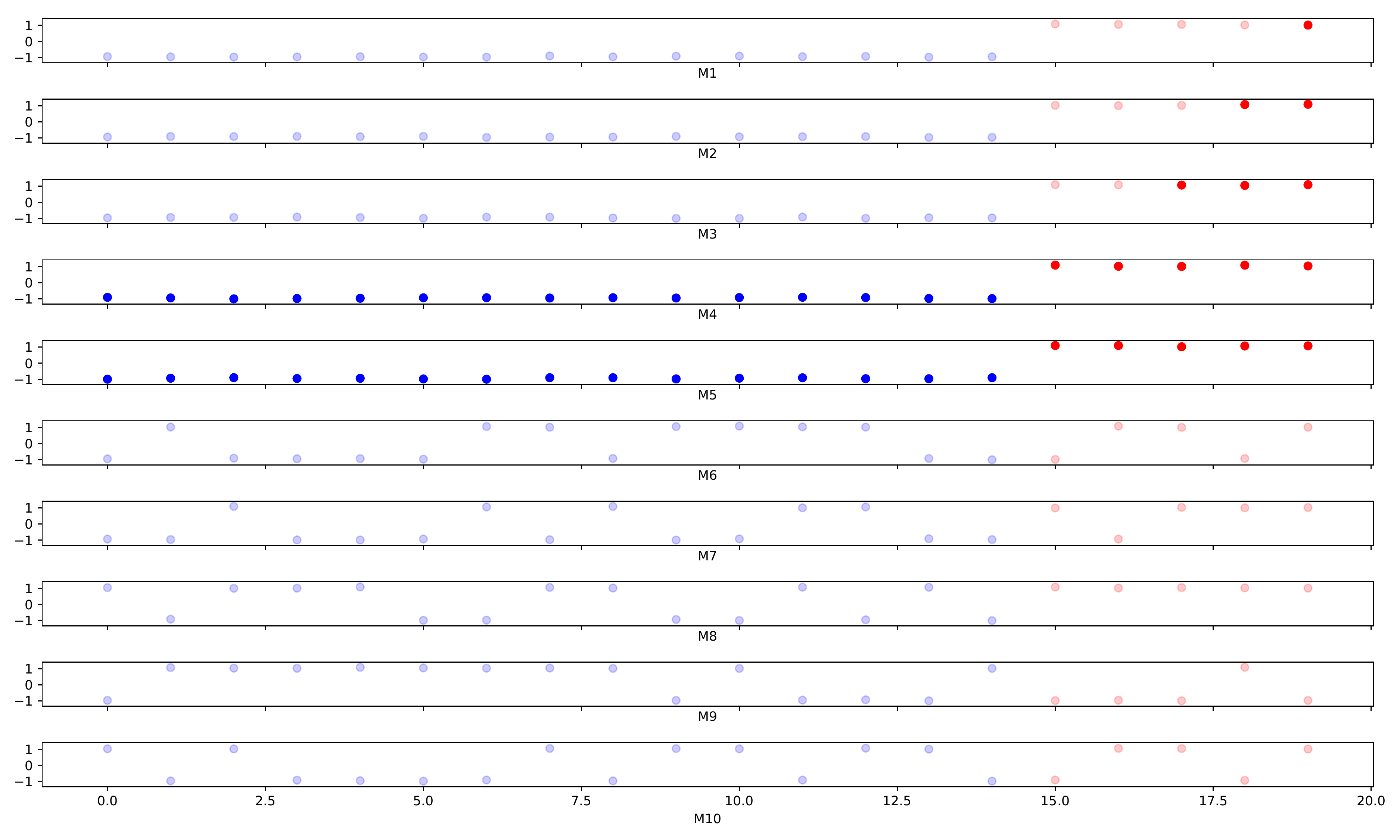}
  \caption{An example trajectory of our dueling DQN policy in the simulation. Blue color denotes all the measurements for healthy state and red for critical state. Darker color represents the measurements taken by the agent. For example, $M4$ and $M5$ are taken all the time to probe the state of the patient.}
  \label{fig:policy_illustration_simulation}
\end{figure}

The goal of this simulation is to study the performance of the RL agent given a near-perfect classifier. Here, we simulate a terminal event forecasting task, use a softmax classifier to produce rewards and then train a Dueling DQN agent for measurement scheduling using the rewards generated by the classifier.

\textbf{Simulation data}
Patient clinical status is simulated to be a binary time series generated under a two-state Markov model: $M = \{0, 1: 0 = Healthy, 1 = Critical\}$. A consecutive sequence of five $1$s in the status series indicates the onset of a terminal event. 
We simulate patients to have different trajectory lengths $T$ indexed by $t$ and $10$ types of input signals indexed by $k$, as follows. 
Let $\epsilon_{t,k} \sim N(0, 0.1)$. The first five types of measurements $(k \in [1,5])$ $y_{t, k} = 1 + \epsilon_{t,k}$ when $S_t=1$ and $-1 + \epsilon_{t,k}$ otherwise. 
The last five types of measurements $(k\in[6,10])$ are $\epsilon_{t,k}$ independent of $S_t$. 
We randomly remove 50\% of the values from the generated matrix to introduce missingness creating a more realistic scenario. In the case of missingness, the measurement value is set to $0$. 
The measurements are designed such that first five types of measurements have increasing importance while the last five measurements are noise.

\textbf{Designed classifier}
We design a classifier considering the feature importance vector $\{f_k\}_{k=1}^{10} = $ (1 2 3 4 5 0 0 0 0 0). 
The classifier takes in measurements of the $5$ most recent timepoints $\{\{y_{t,k}\}_{k=1}^{10}\}_{t=t'-4}^{t'}$, where $t'$ is the current time. 
Let $\eta$ denote a time decay factor, where past measurements are less important. The classifier then forecasts whether the patient experiences a  terminal event within 5 future timepoints with $p(o_{t'+5}=1) = softmax( \sum_{t=t'-4}^{t'} \sum_{k=1}^{10} y_{t, k} \cdot f_k \cdot \eta^t)$. 
The classifier increases the certainty of a terminal event when it discovers more critical signals in the measurement values. 
To see whether the agent can distinguish features with different importance, we employ a uniform action cost $c(a_v) = 1$.
The RL agent takes $\{\{y_{t,k}\}_{k=1}^{10}\}_{t=t'-4}^{t'}$ as input. We set reward discount factor $\gamma = 0.999$ in this task. 

We simulated a dataset of $5,000$ patient trajectories with $T \in [25, 50]$ according to the scheme above. 5\% of the patients end up with a terminal event. We learn several dueling DQN agents by varying trade-off factor $\lambda$. We include several baselines that resemble the heuristic-based test scheduling. One of the baseline policies is randomly selecting $x$ informative measurements ($Random\_informative$), where $x \in [1, 5]$. Another class of baseline policies is randomly selecting $x$ measurements ($Random$), where $x \in [1, 10]$.

As we vary $\lambda$, we learn a range of policies that trade off between action frequency and predictive probability of detecting the terminal event (Figure \ref{fig:reward_roc_plot_simulation}). Under the same action frequency, our learned dueling DQN agent consistently outperform baseline policies in terms of predictive probability of detecting disease, showing the benefits of dynamically measure patients conditioned on the patient state.

We show an example patient trajectory of our dueling DQN policy in Figure \ref{fig:policy_illustration_simulation}. It always selects the most and the second most informative features ($M4$, $M5$) to probe which state the patient is in. It sequentially selects the other informative features ($M3$, $M2$, $M1$) whenever it finds the patient is in a critical state. It doesn't select any noisy features to avoid accruing total measurement cost.


\subsection{Results on MIMIC3}
\label{sec:mimic}

Here we test our policy on a real-world ICU dataset MIMIC3 to gain better clinical sampling policy.
The details of our preprocessing of MIMIC3 are in the Appendix \ref{appendix:preprocessing}.
First, we train a mortality forecasting model. 
Our task is to predict if patient dies within $24$ hours given the past 24 hours of observations. The observations include $39$ time-series measurements and $38$ static covariates.

\begin{table}[t]
  \centering 
  \caption{The test set performances of the trained classifiers in $24$ hour mortality prediction.} 
\begin{tabular}{|l|l|l|}
  \toprule
& AUC & AUPR \\ \midrule
LR & $0.931$ & $0.752$ \\
RF & $0.935$ & $0.756$ \\ \midrule
RNN & $0.950$ & $0.803$ \\ 
\bottomrule
\end{tabular}
\label{table:classifier_performance}
\end{table}

We show that we train a well performing RNN classifier: with sufficient information RNN vastly outperforms baselines such as random forest that do not consider long-term dependency (Table \ref{table:classifier_performance}). 
The details of the classifier training are in Appendix \ref{appendix:classifier_performance}. 
By combining the classifier and RL, we are able to learn clinically relevant policies from off-line data and show our policies perform better than clinician's policy using off-policy policy evaluation.

\paragraph{Training policies and off-policy evaluation}
We take each patient's last $24$ hours and discretize the experience into $30$-minutes intervals, leading to $48$ time points.
We remove the patients with fewer than $12$ hours of recording or less than $5$ measurements available.
We set $\gamma = 0.95$ to encourage the agent to increase predictive performance earlier rather than later.
We vary our DQN architecture, random seed and action cost coefficient $\lambda$ (range listed in Table \ref{table:dqn_hyperparameters}) to train $150$ different policies by random search, and select the best performing policies based on validation set.
We list all the hyperparameters in Appendix Table \ref{table:dqn_hyperparameters}.

\begin{figure}[t]
  \centering
  \includegraphics[width=0.5\textwidth]{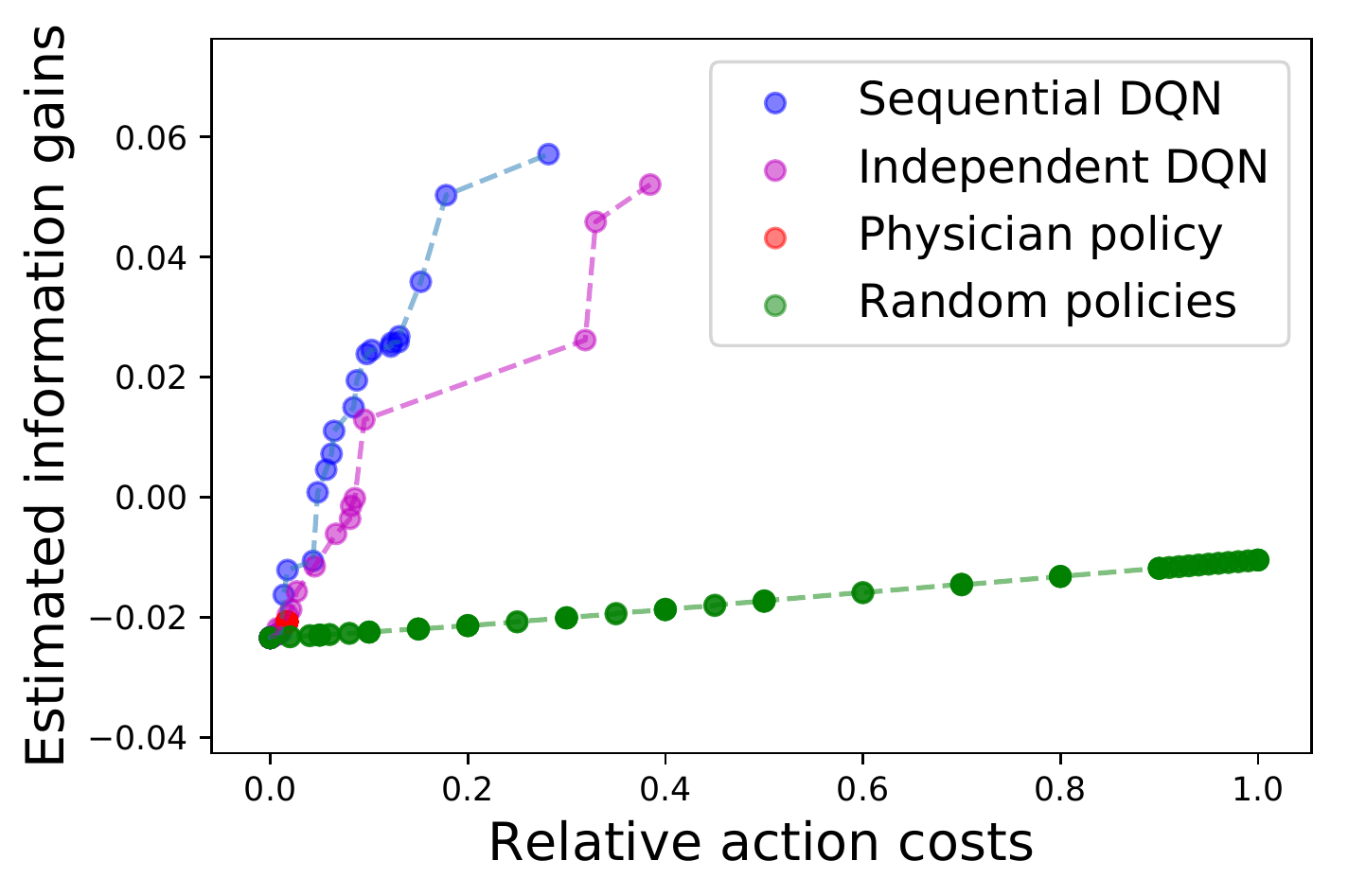}
  \caption{Offline evaluation of physician, sequential DQN, independent DQN and random policies in MIMIC3. Relative action cost is normalized between $0$ and $1$ for the accumulated action costs $c$. }
  \label{fig:reward_roc_mimic_full}
\end{figure}

\begin{figure}[t]
  \centering
  \includegraphics[width=0.5\textwidth]{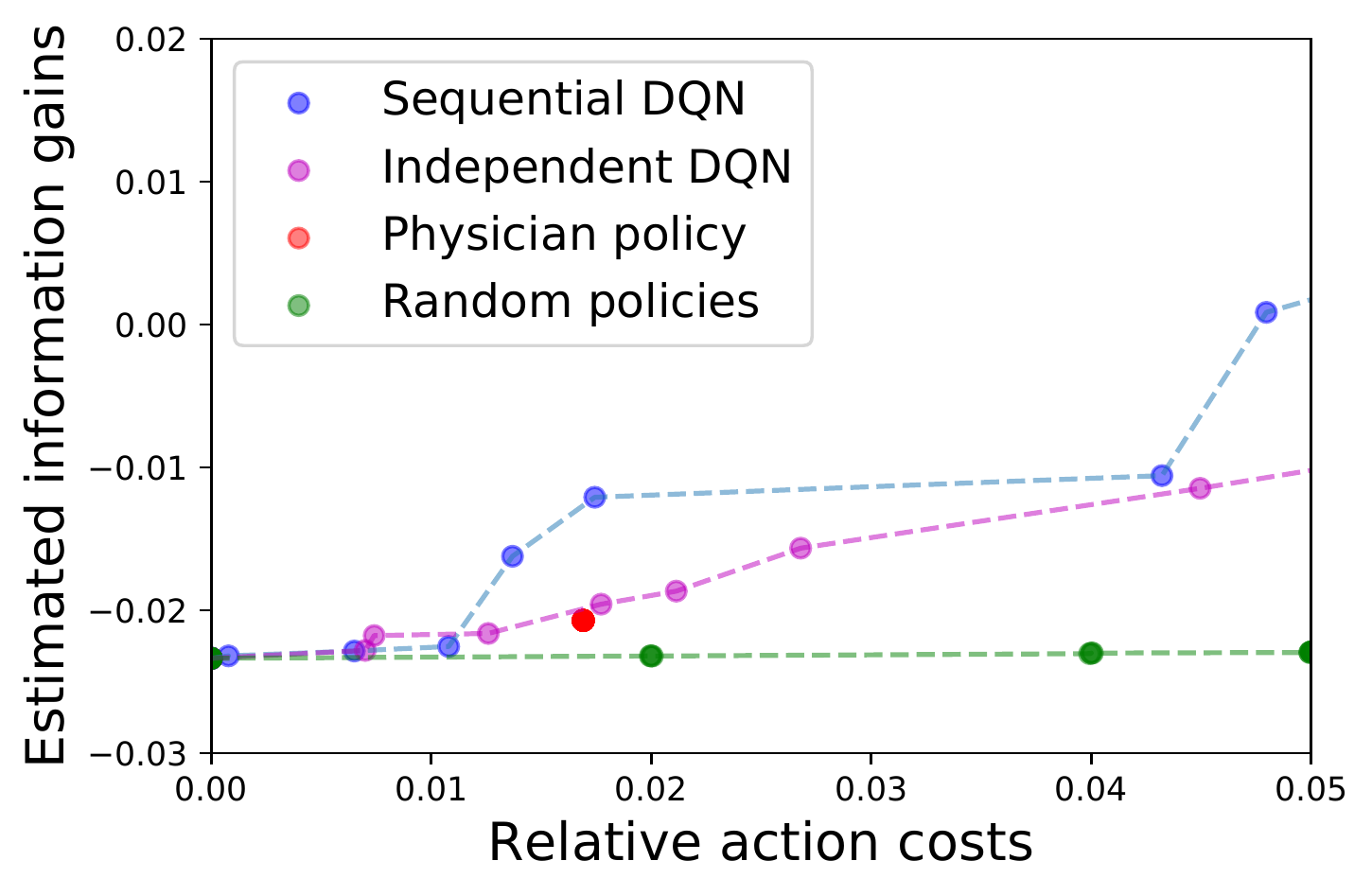}
  \caption{Focus view of Figure \ref{fig:reward_roc_mimic_full}. Compared to physician policy, our policies reduce around $31\%$ of the action costs under the same information gain, or $3$ times increase of the information gain under the same action costs relative to the lowest information gain. }
  \label{fig:reward_roc_mimic_focus}
\end{figure}

We use regression based OPPE to evaluate our agent policies, physician policy and random policies shown in Figure \ref{fig:reward_roc_mimic_full}. 
Ideally, a great policy should have low action frequency and high information gain.
By interpolating across various DQN performing points, we can get a frontier of performance curve for our DQN agent.
Using this frontier, compared to physician policy, our policies (denoted sequential DQN in Figure \ref{fig:reward_roc_mimic_focus}) reduce action costs by $31\%$ under the same information gain, or increase the information gain $3$ times relative to the lowest information gain with the same action costs.
In addition, we scale up \citet{Cheng2018} to our setting by considering the reward independently for each measurement and model the Q-values using a multi-output neural network (denoted independent DQN in Figure \ref{fig:reward_roc_mimic_full}). This approach only increases information gain by 30\%, decreasing the cost by 12\%.

The lowest information gain is the information gain when no measurements are taken. 
Maybe surprisingly, sampling all the measurements (relative action cost as $1$) do not produce the policy with the highest information gain.
We think it is because the classifier tends to make mistakes on some measurements so measuring everything decreases the classifier's performance. Or it could be the measurement itself is rarely measured or noisy and that confuses the classifier.

\begin{figure*}[t]
  \centering
  \includegraphics[width=\textwidth]{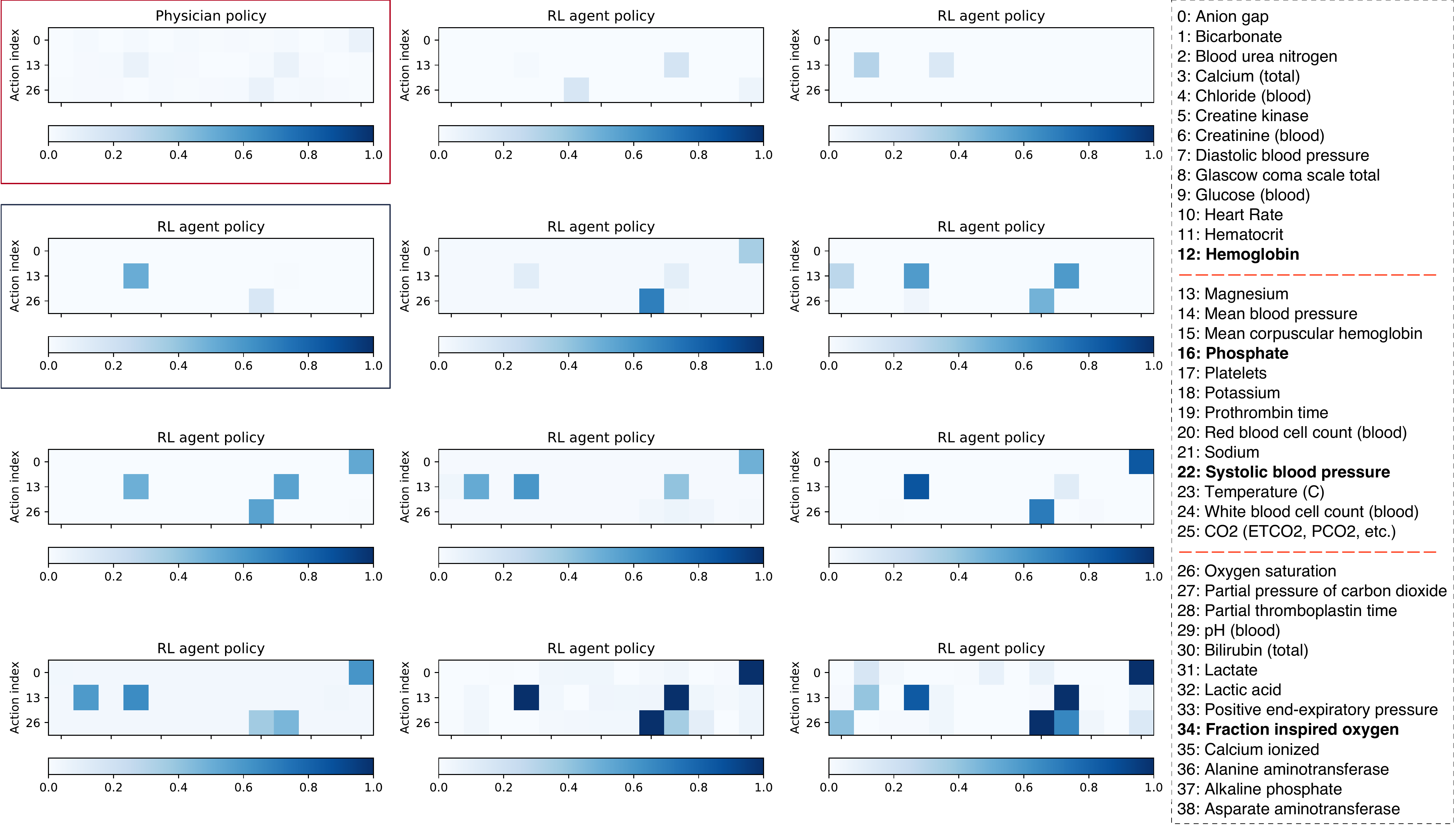}
  \caption{
  Action frequency of physician and RL agent policies. 
  Each panel has 39 cells aligned on three rows with each row of 13 actions (e.g. 1st row for action 0-12 etc.), with color representing the sampling frequency per action. The measurement corresponding to the action index is shown on the right. 
  Top left is the action frequency of physician policy in our 24h mortality forecasting task. The rest are the action frequencies of Sequential DQN policies with increasing action costs, from left to right, from top to bottom. 
  Physician's most selected measurements in the last 24h are highlighted in bold. The RL agent policy with comparable action cost as physician policy but higher information gains is highlighted in the black box. 
  }
  \label{fig:policy_illustration_mimic}
\end{figure*}

We compare our policies' action frequency with physician's action frequency to gain insights from the learned policies (Figure \ref{fig:policy_illustration_mimic}).
We show our policies with increasing action frequency, from left to right, top to bottom.
The most frequent measurements performed by physicians within the last $24$ hours (red box) are Hemoglobins, Phosphate, Systolic blood pressure and Fraction inspired oxygen (FiO2) (see Appendix Table \ref{table:time_var_features_24h_mortality} for a full list), indicating the clinical importance of these $4$ measurements.
It is reassuring that the closest policy with the same action costs (black box) also focus on these $4$ most frequent measurements with focus on Phosphate and FiO2. We find these $2$ are strongly correlated with the death in ICU due to Hypophosphatemia \citep{geerse2010treatment,miller2018impact} and serving as important functional indicators or hyperoxia \citep{damiani2014arterial,ramanan2018association}. 
As we increase measurement costs, our policies select other features like Calcium Ionized, Mean blood pressure and Oxygen Saturation, indicating the importance of these features for the task of mortality prediction.



\begin{figure}[t]
  \centering
  \includegraphics[width=0.5\textwidth]{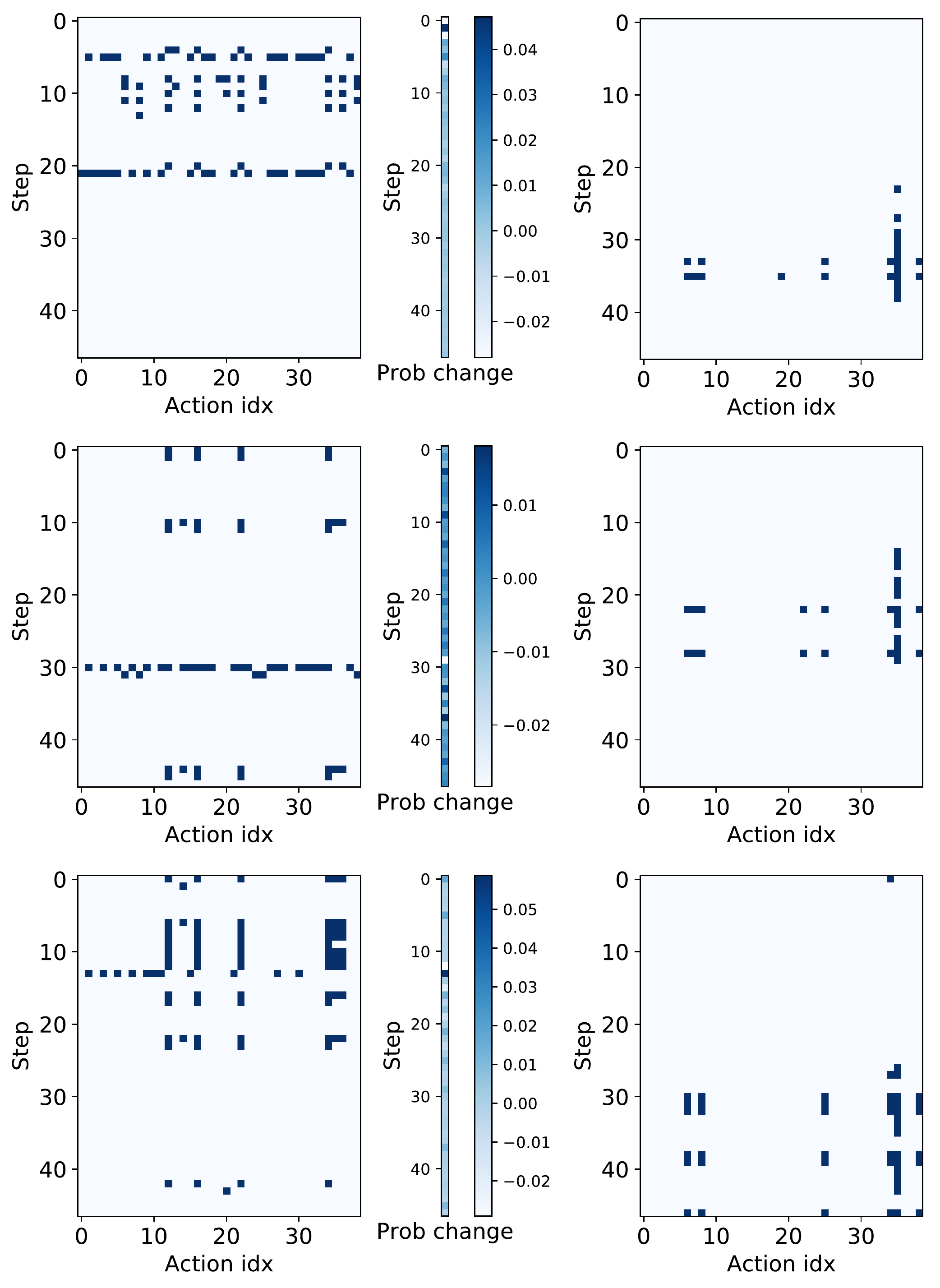}
  \caption{Examples of sampling strategy in the last $24$ hours of $3$ dying patients. (Left) Physician policy. (Middle) Probability change due to physician' measurement. (Right) Actions performed by sequential DQN. Sequential DQN makes decision based on the history of the physician's sampled measurements and it is adaptive to patients’ history, recommending probing patients when no recent measurement exists and vice versa.
}
  \label{fig:policy_trajectory_mimic}
\end{figure}

In Figure \ref{fig:policy_trajectory_mimic}, we compare our agent's and physician's sampling strategy for the last $24$ hours of $3$ dying patients. 
Our agent starts scheduling measurements when there is no recent measurements made and stops when the measurements are taken recently.
This shows that our policies dynamically sample adaptive to the patient's condition, rather than a simple rule-of-thumb measurement strategy. 

\section{Discussion and Future Work}
In this work, we propose a scalable method for measurement scheduling using data-driven approach. 
We show that our scheduling policy achieves better predictive power with lower measurement costs in both simulations and on MIMIC3. 
We also examine our policies qualitatively and show that our policies sample clinically-relevant measurements and act based on the patient's measurement history.

In this study, we assume sampling lab tests at the current time point provide us with information by the next time point. In reality, some lab tests can take hours to get the information back. 
In our framework, we can incorporate the time constraint and relax this assumption by delaying the reward that RL agent receives to a later timepoint to adjust for this bias. 
We also assume all medical measurements are needed for scheduling and we only consider the simplest setting that all measurements have the same cost. 
But it is true that some routines or low-cost measurements do not have to be considered for scheduling, and some measurements can be grouped by using prior clinical knowledge.
Also, we did not incorporate treatment information which can be valuable for improving classifier performance. 
Recent literature on incorporating treatment information to model physiologic signals using causal inference can help address this issue \citep{schulam2017if, soleimani2017treatment}. 

While in this work we only learn from one fixed classifier, dataset shift or covariate shift problems can arise due to change in scheduling policy. 
One way to solve this is to sequentially train the classifier on newly collected data and retrain RL agent on the classifier. 
Another concern is that our RL agent might learn to sample features that are over-fitting to the classifier. 
To avoid this phenomenon, we plan to scale our work by using an ensemble of classifiers to reduce over-fitting and produce a robust reward. 

Using regression-based value estimator is known to have provably low variance when the MDP is well estimated. However, bias can be introduced since real-world problems usually have a large state space, and many state-action pairs will not be observed in the data. 
We plan to use more complex model for the regression based value estimator or use better value estimator \citep{Liu2018RepresentationBM} to capture the underlying dynamics. Further, incorporating causal thinking in RL framework might help learn safer policies, for example, recent work by \citet{kallus2018confounding} presents a model for personalized decision policy learning in the presence of unobserved confounding and its application to acute ischaemic stroke treatment.

Besides determining what lab tests could improve the early warning system, it would also be interesting to see what lab tests could help make the right medical decisions. For example, what measurements to schedule that leads to a drug administration or a treatment procedure. We leave this for future work.

We will also investigate interpretability for our RL policies, motivated by saliency maps \citep{chang2018explaining} and example-based methods \citep{joshi2018xgems}, helping to provide clinical insights.
 
We believe that RL has a great role to play in helping diagnose and ultimately prevent critical events in the healthcare.

\subsubsection*{Acknowledgments}
We thank Amir-massoud Farahmand and Marzyeh Ghassemi for their helpful discussions. AG and CC are supported by the Early Researcher Award from the Ministry of Research and Innovation.

\bibliography{example_paper}

\begin{thebibliography}{43}
\providecommand{\natexlab}[1]{#1}
\providecommand{\url}[1]{\texttt{#1}}
\expandafter\ifx\csname urlstyle\endcsname\relax
  \providecommand{\doi}[1]{doi: #1}\else
  \providecommand{\doi}{doi: \begingroup \urlstyle{rm}\Url}\fi

\bibitem[Ahuja et~al.(2017)Ahuja, Zame, and van~der
  Schaar]{ahuja_dpscreen_2017}
Ahuja, K., Zame, W., and van~der Schaar, M.
\newblock {DPSCREEN}: {Dynamic} {Personalized} {Screening}.
\newblock In \emph{Advances in {Neural} {Information} {Processing} {Systems}},
  pp.\  1321--1332, 2017.

\bibitem[Brodersen et~al.(2018)Brodersen, Schwartz, Heneghan, O’Sullivan,
  Aronson, and Woloshin]{brodersen2018overdiagnosis}
Brodersen, J., Schwartz, L.~M., Heneghan, C., O’Sullivan, J.~W., Aronson,
  J.~K., and Woloshin, S.
\newblock Overdiagnosis: what it is and what it isn’t, 2018.

\bibitem[Chang et~al.(2019)Chang, Creager, Goldenberg, and
  Duvenaud]{chang2018explaining}
Chang, C.-H., Creager, E., Goldenberg, A., and Duvenaud, D.
\newblock Explaining image classifiers by counterfactual generation.
\newblock In \emph{International Conference on Learning Representations}, 2019.
\newblock URL \url{https://openreview.net/forum?id=B1MXz20cYQ}.

\bibitem[Cheng et~al.(2019)Cheng, Prasad, and Engelhardt]{Cheng2018}
Cheng, L.-F., Prasad, N., and Engelhardt, B.~E.
\newblock {An optimal policy for patient laboratory tests in intensive care
  units}.
\newblock \emph{Proceedings of the Pacific Symposium on Biocomputing (PSB)},
  2019.
\newblock URL
  \url{https://psb.stanford.edu/psb-online/proceedings/psb19/cheng_l.pdf}.

\bibitem[Contardo et~al.(2016)Contardo, Denoyer, and
  Arti{\`e}res]{contardo2016sequential}
Contardo, G., Denoyer, L., and Arti{\`e}res, T.
\newblock Sequential cost-sensitive feature acquisition.
\newblock In \emph{International Symposium on Intelligent Data Analysis}, pp.\
  284--294. Springer, 2016.

\bibitem[Damiani et~al.(2014)Damiani, Adrario, Girardis, Romano, Pelaia,
  Singer, and Donati]{damiani2014arterial}
Damiani, E., Adrario, E., Girardis, M., Romano, R., Pelaia, P., Singer, M., and
  Donati, A.
\newblock Arterial hyperoxia and mortality in critically ill patients: a
  systematic review and meta-analysis.
\newblock \emph{Critical Care}, 18\penalty0 (6):\penalty0 711, 2014.

\bibitem[Dewan et~al.(2017)Dewan, Galvez, Polsky, Kreher, Kraus, Ahumada,
  McCloskey, and Wolfe]{dewan2017reducing}
Dewan, M., Galvez, J., Polsky, T., Kreher, G., Kraus, B., Ahumada, L.,
  McCloskey, J., and Wolfe, H.
\newblock Reducing unnecessary postoperative complete blood count testing in
  the pediatric intensive care unit.
\newblock \emph{The Permanente journal}, 21, 2017.

\bibitem[Feldman(2009)]{feldman2009managing}
Feldman, L.
\newblock Managing the cost of diagnosis.
\newblock \emph{Managed care (Langhorne, Pa.)}, 18\penalty0 (5):\penalty0 43,
  2009.

\bibitem[Futoma et~al.(2017)Futoma, Hariharan, Sendak, Brajer, Clement, Bedoya,
  O'Brien, and Heller]{futoma_improved_2017}
Futoma, J., Hariharan, S., Sendak, M., Brajer, N., Clement, M., Bedoya, A.,
  O'Brien, C., and Heller, K.
\newblock An {Improved} {Multi}-{Output} {Gaussian} {Process} {RNN} with
  {Real}-{Time} {Validation} for {Early} {Sepsis} {Detection}.
\newblock \emph{arXiv:1708.05894 [stat]}, August 2017.
\newblock arXiv: 1708.05894.

\bibitem[Futoma et~al.(2018)Futoma, Lin, Sendak, Bedoya, Clement, O'Brien, and
  Heller]{futoma2018learning}
Futoma, J., Lin, A., Sendak, M., Bedoya, A., Clement, M., O'Brien, C., and
  Heller, K.
\newblock Learning to treat sepsis with multi-output gaussian process deep
  recurrent q-networks, 2018.
\newblock URL \url{https://openreview.net/forum?id=SyxCqGbRZ}.

\bibitem[Geerse et~al.(2010)Geerse, Bindels, Kuiper, Roos, Spronk, and
  Schultz]{geerse2010treatment}
Geerse, D.~A., Bindels, A.~J., Kuiper, M.~A., Roos, A.~N., Spronk, P.~E., and
  Schultz, M.~J.
\newblock Treatment of hypophosphatemia in the intensive care unit: a review.
\newblock \emph{Critical Care}, 14\penalty0 (4):\penalty0 R147, 2010.

\bibitem[Ghassemi et~al.(2015)Ghassemi, Pimentel, Naumann, Brennan, Clifton,
  Szolovits, and Feng]{ghassemi2015multivariate}
Ghassemi, M., Pimentel, M.~A., Naumann, T., Brennan, T., Clifton, D.~A.,
  Szolovits, P., and Feng, M.
\newblock A multivariate timeseries modeling approach to severity of illness
  assessment and forecasting in icu with sparse, heterogeneous clinical data.
\newblock 2015.

\bibitem[Harutyunyan et~al.(2017)Harutyunyan, Khachatrian, Kale, and
  Galstyan]{harutyunyan2017multitask}
Harutyunyan, H., Khachatrian, H., Kale, D.~C., and Galstyan, A.
\newblock Multitask learning and benchmarking with clinical time series data.
\newblock \emph{arXiv preprint arXiv:1703.07771}, 2017.

\bibitem[He et~al.(2016)He, Mineiro, and Karampatziakis]{he2016active}
He, H., Mineiro, P., and Karampatziakis, N.
\newblock Active information acquisition.
\newblock \emph{arXiv preprint arXiv:1602.02181}, 2016.

\bibitem[Hochreiter \& Schmidhuber(1997)Hochreiter and
  Schmidhuber]{hochreiter1997long}
Hochreiter, S. and Schmidhuber, J.
\newblock Long short-term memory.
\newblock \emph{Neural computation}, 9\penalty0 (8):\penalty0 1735--1780, 1997.

\bibitem[Hoffman \& Cooper(2012)Hoffman and Cooper]{hoffman2012overdiagnosis}
Hoffman, J.~R. and Cooper, R.~J.
\newblock Overdiagnosis of disease: a modern epidemic.
\newblock \emph{Archives of internal medicine}, 172\penalty0 (15):\penalty0
  1123--1124, 2012.

\bibitem[Iosfina et~al.(2013)Iosfina, Merkeley, Cessford, Geller, Amiri,
  Baradaran, Norena, Ayas, and Dodek]{iosfina2013implementation}
Iosfina, I., Merkeley, H., Cessford, T., Geller, G., Amiri, N., Baradaran, N.,
  Norena, M., Ayas, N., and Dodek, P.~M.
\newblock Implementation of an on-demand strategy for routine blood testing in
  icu patients.
\newblock In \emph{D23. QUALITY IMPROVEMENT IN CRITICAL CARE}, pp.\
  A5322--A5322. Am Thoracic Soc, 2013.

\bibitem[Jiang \& Li(2015)Jiang and Li]{jiang2015doubly}
Jiang, N. and Li, L.
\newblock Doubly robust off-policy value evaluation for reinforcement learning.
\newblock \emph{arXiv preprint arXiv:1511.03722}, 2015.

\bibitem[Johnson et~al.(2016)Johnson, Pollard, Shen, Li-wei, Feng, Ghassemi,
  Moody, Szolovits, Celi, and Mark]{johnson2016mimic}
Johnson, A.~E., Pollard, T.~J., Shen, L., Li-wei, H.~L., Feng, M., Ghassemi,
  M., Moody, B., Szolovits, P., Celi, L.~A., and Mark, R.~G.
\newblock Mimic-iii, a freely accessible critical care database.
\newblock \emph{Scientific data}, 3:\penalty0 160035, 2016.

\bibitem[Joshi et~al.(2018)Joshi, Koyejo, Kim, and Ghosh]{joshi2018xgems}
Joshi, S., Koyejo, O., Kim, B., and Ghosh, J.
\newblock xgems: Generating examplars to explain black-box models.
\newblock \emph{arXiv preprint arXiv:1806.08867}, 2018.

\bibitem[Kallus \& Zhou(2018)Kallus and Zhou]{kallus2018confounding}
Kallus, N. and Zhou, A.
\newblock Confounding-robust policy improvement.
\newblock \emph{arXiv preprint arXiv:1805.08593}, 2018.

\bibitem[Komorowski et~al.(2018)Komorowski, Celi, Badawi, Gordon, and
  Faisal]{komorowski2018artificial}
Komorowski, M., Celi, L.~A., Badawi, O., Gordon, A.~C., and Faisal, A.~A.
\newblock The artificial intelligence clinician learns optimal treatment
  strategies for sepsis in intensive care.
\newblock \emph{Nature Medicine}, 24\penalty0 (11):\penalty0 1716, 2018.

\bibitem[Kotecha et~al.(2017)Kotecha, Shapiro, Cardasis, and
  Narayanswami]{kotecha2017reducing}
Kotecha, N., Shapiro, J.~M., Cardasis, J., and Narayanswami, G.
\newblock Reducing unnecessary laboratory testing in the medical icu.
\newblock \emph{The American journal of medicine}, 130\penalty0 (6):\penalty0
  648--651, 2017.

\bibitem[Li \& Marlin(2016)Li and Marlin]{li2016scalable}
Li, S. C.-X. and Marlin, B.~M.
\newblock A scalable end-to-end gaussian process adapter for irregularly
  sampled time series classification.
\newblock In \emph{Advances in neural information processing systems}, pp.\
  1804--1812, 2016.

\bibitem[Lipton et~al.()Lipton, Kale, and Wetzel]{lipton2016modeling}
Lipton, Z.~C., Kale, D.~C., and Wetzel, R.
\newblock Modeling missing data in clinical time series with rnns.

\bibitem[Liu et~al.(2018)Liu, Gottesman, Raghu, Komorowski, Faisal,
  Doshi-Velez, and Brunskill]{Liu2018RepresentationBM}
Liu, Y., Gottesman, O., Raghu, A., Komorowski, M., Faisal, A., Doshi-Velez, F.,
  and Brunskill, E.
\newblock Representation balancing mdps for off-policy policy evaluation.
\newblock \emph{CoRR}, abs/1805.09044, 2018.

\bibitem[Miller et~al.()Miller, Doepker, Springer, Exline, Phillips, and
  Murphy]{miller2018impact}
Miller, C.~J., Doepker, B.~A., Springer, A.~N., Exline, M.~C., Phillips, G.,
  and Murphy, C.~V.
\newblock Impact of serum phosphate in mechanically ventilated patients with
  severe sepsis and septic shock.
\newblock \emph{Journal of intensive care medicine}, pp.\  0885066618762753.

\bibitem[Pageler et~al.(2013)Pageler, Franzon, Longhurst, Wood, Shin, Adams,
  Widen, and Cornfield]{pageler2013embedding}
Pageler, N.~M., Franzon, D., Longhurst, C.~A., Wood, M., Shin, A.~Y., Adams,
  E.~S., Widen, E., and Cornfield, D.~N.
\newblock Embedding time-limited laboratory orders within computerized provider
  order entry reduces laboratory utilization.
\newblock \emph{Pediatric Critical Care Medicine}, 14\penalty0 (4):\penalty0
  413--419, 2013.

\bibitem[Paxton et~al.(2013)Paxton, Niculescu-Mizil, and
  Saria]{paxton2013developing}
Paxton, C., Niculescu-Mizil, A., and Saria, S.
\newblock Developing predictive models using electronic medical records:
  challenges and pitfalls.
\newblock In \emph{AMIA Annual Symposium Proceedings}, volume 2013, pp.\  1109.
  American Medical Informatics Association, 2013.

\bibitem[Prasad et~al.(2017)Prasad, Cheng, Chivers, Draugelis, and
  Engelhardt]{prasad2017reinforcement}
Prasad, N., Cheng, L.-F., Chivers, C., Draugelis, M., and Engelhardt, B.~E.
\newblock A reinforcement learning approach to weaning of mechanical
  ventilation in intensive care units.
\newblock \emph{arXiv preprint arXiv:1704.06300}, 2017.

\bibitem[Raghu et~al.(2017)Raghu, Komorowski, Celi, Szolovits, and
  Ghassemi]{raghu2017continuous}
Raghu, A., Komorowski, M., Celi, L.~A., Szolovits, P., and Ghassemi, M.
\newblock Continuous state-space models for optimal sepsis treatment-a deep
  reinforcement learning approach.
\newblock \emph{arXiv preprint arXiv:1705.08422}, 2017.

\bibitem[Ramanan \& Fisher(2018)Ramanan and Fisher]{ramanan2018association}
Ramanan, M. and Fisher, N.
\newblock The association between arterial oxygen tension, hemoglobin
  concentration, and mortality in mechanically ventilated critically ill
  patients.
\newblock \emph{Indian journal of critical care medicine: peer-reviewed,
  official publication of Indian Society of Critical Care Medicine},
  22\penalty0 (7):\penalty0 477, 2018.

\bibitem[Salisbury et~al.(2011)Salisbury, Reid, Alexander, Masoudi, Lai, Chan,
  Bach, Wang, Spertus, and Kosiborod]{salisbury2011diagnostic}
Salisbury, A.~C., Reid, K.~J., Alexander, K.~P., Masoudi, F.~A., Lai, S.-M.,
  Chan, P.~S., Bach, R.~G., Wang, T.~Y., Spertus, J.~A., and Kosiborod, M.
\newblock Diagnostic blood loss from phlebotomy and hospital-acquired anemia
  during acute myocardial infarction.
\newblock \emph{Archives of internal medicine}, 171\penalty0 (18):\penalty0
  1646--1653, 2011.

\bibitem[Schulam \& Saria(2017)Schulam and Saria]{schulam2017if}
Schulam, P. and Saria, S.
\newblock What-if reasoning with counterfactual gaussian processes.
\newblock \emph{History}, 100:\penalty0 120, 2017.

\bibitem[Shim et~al.(2017)Shim, Hwang, and Yang]{shim2017pay}
Shim, H., Hwang, S.~J., and Yang, E.
\newblock Why pay more when you can pay less: A joint learning framework for
  active feature acquisition and classification.
\newblock \emph{arXiv preprint arXiv:1709.05964}, 2017.

\bibitem[Soleimani et~al.(2017{\natexlab{a}})Soleimani, Hensman, and
  Saria]{soleimani_scalable_2017}
Soleimani, H., Hensman, J., and Saria, S.
\newblock Scalable {Joint} {Models} for {Reliable} {Uncertainty}-{Aware}
  {Event} {Prediction}.
\newblock \emph{arXiv:1708.04757 [cs, stat]}, August 2017{\natexlab{a}}.
\newblock URL \url{http://arxiv.org/abs/1708.04757}.
\newblock 00000 arXiv: 1708.04757.

\bibitem[Soleimani et~al.(2017{\natexlab{b}})Soleimani, Subbaswamy, and
  Saria]{soleimani2017treatment}
Soleimani, H., Subbaswamy, A., and Saria, S.
\newblock Treatment-response models for counterfactual reasoning with
  continuous-time, continuous-valued interventions.
\newblock \emph{arXiv preprint arXiv:1704.02038}, 2017{\natexlab{b}}.

\bibitem[Soleimani et~al.(2018)Soleimani, Hensman, and
  Saria]{soleimani2018scalable}
Soleimani, H., Hensman, J., and Saria, S.
\newblock Scalable joint models for reliable uncertainty-aware event
  prediction.
\newblock \emph{IEEE transactions on pattern analysis and machine
  intelligence}, 40\penalty0 (8):\penalty0 1948--1963, 2018.

\bibitem[Wang et~al.(2018)Wang, Zhang, He, and Zha]{wang2018supervised}
Wang, L., Zhang, W., He, X., and Zha, H.
\newblock Supervised reinforcement learning with recurrent neural network for
  dynamic treatment recommendation.
\newblock In \emph{Proceedings of the 24th ACM SIGKDD International Conference
  on Knowledge Discovery \& Data Mining}, pp.\  2447--2456. ACM, 2018.

\bibitem[Wang et~al.(2015)Wang, Schaul, Hessel, Van~Hasselt, Lanctot, and
  De~Freitas]{wang2015dueling}
Wang, Z., Schaul, T., Hessel, M., Van~Hasselt, H., Lanctot, M., and De~Freitas,
  N.
\newblock Dueling network architectures for deep reinforcement learning.
\newblock \emph{arXiv preprint arXiv:1511.06581}, 2015.

\bibitem[Weng et~al.(2017)Weng, Gao, He, Yan, and
  Szolovits]{weng2017representation}
Weng, W.-H., Gao, M., He, Z., Yan, S., and Szolovits, P.
\newblock Representation and reinforcement learning for personalized glycemic
  control in septic patients.
\newblock \emph{arXiv preprint arXiv:1712.00654}, 2017.

\bibitem[Yoon et~al.(2018)Yoon, Zame, and van~der Schaar]{yoon2018deep}
Yoon, J., Zame, W.~R., and van~der Schaar, M.
\newblock Deep sensing: Active sensing using multi-directional recurrent neural
  networks.
\newblock 2018.

\bibitem[Zhang et~al.(2017)Zhang, Xie, Wang, and Xing]{zhang2017medical}
Zhang, S., Xie, P., Wang, D., and Xing, E.~P.
\newblock Medical diagnosis from laboratory tests by combining generative and
  discriminative learning.
\newblock \emph{arXiv preprint arXiv:1711.04329}, 2017.

\end{thebibliography}
\bibliographystyle{icml2019}

\clearpage
\appendix



\section{MIMIC3 Preprocessing for Survival Forecasting}
\label{appendix:preprocessing}
We use the publicly available dataset MIMIC3 \citep{johnson2016mimic} and then follow the preprocessing of \citet{harutyunyan2017multitask} for the in-hospital mortality prediction task. It excludes the neonatal and pediatric patients 
and patients with multiple ICU stays. The training set consists of $35,725$ patients with $10.81\%$ mortality rate, and test set has $6,294$ patients with $9.94\%$ mortality rate.
We then split $15\%$ of our training set as our validation set. 



For classifier training, we uniformly take $6$ timepoints within the last $24$ hours of each patient trajectory. For each prediction point, we set the label as $1$ if the the patient die in the encounter and $0$ otherwise. 
For each patient, we evaluate on a uniformly spaced grid points with separation of 3 hours starting backward from the patient’s last time until the maximum 24 hours, resulting in maximum 7 points per patient.
We only include the prediction points with at least $3$ hours of history and $5$ measurement values. The ultimate goal is to predict whether patient dies within 24 hours given the past 24 hours of observations.

For RL, we take the last $24$ hours of each dying patient and discretize it into $30$ minutes interval. 
We only include the patients with at least $12$ hours to remove unstable trajectories.
Note that we set the RL trajectory and classifier prediction horizon as the last $24$ hours of the patient.
This could avoid the label confounding problem \citep{paxton2013developing} as there might be unrecorded intervention that increase patient's health condition and change the label.

  
  
We select $38$ static demographic features and clinical.
Age, Gender, Ethnicity, congestive heart failure, cardiac arrhythmias, valvular disease, pulmonary circulation, peripheral vascular, hypertension, paralysis, other neurological, chronic pulmonary, diabetes uncomplicated, diabetes complicated, hypothyroidism, renal failure, liver disease, peptic ulcer, aids, lymphoma, metastatic cancer, solid tumor, rheumatoid arthritis, coagulopathy, obesity, weight loss, fluid electrolyte, blood loss anemia, deficiency anemias, alcohol abuse, drug abuse, psychoses, depression. The features are curated from the official MIMIC repository \footnote{\url{https://github.com/MIT-LCP/mimic-code/tree/master/concepts}} with the comorbidity concept.

We show the feature choices and their count in Appendix Table \ref{table:time_var_features}.
We select $39$ time-series measurements with counts at least $1\%$ of the count of heart rate, which is the largest count of the measurement in our data. We further log-transform, remove outliers outside of $2$ inter-quantile regions (IRQ), and standardize time series measurement values for zero-mean unit-variance for each feature. 


\begin{table*}[t]
  \centering 
  \caption{Time variant features and their counts after preprocessing} 
\begin{tabular}{l|l|l}
  \toprule
Feature & Count & Relative Count \% \\ \midrule
Anion gap & 213442 & 0.051 \\
Bicarbonate & 219802 & 0.052 \\
Blood urea nitrogen & 220854 & 0.053 \\
Calcium (total) & 185718 & 0.044 \\
Chloride (blood) & 225476 & 0.054 \\
Creatine kinase & 44459 & 0.011 \\
Creatinine (blood) & 221715 & 0.053 \\
Diastolic blood pressure & 3929745 & 0.935 \\
Glascow coma scale total & 627577 & 0.149 \\
Glucose (blood) & 313798 & 0.075 \\
Heart Rate & 4204926 & 1.0 \\
Hematocrit & 253045 & 0.06 \\
Hemoglobin & 196859 & 0.047 \\
Magnesium & 218030 & 0.052 \\
Mean blood pressure & 3904218 & 0.928 \\
Mean corpuscular hemoglobin & 194995 & 0.046 \\
Phosphate & 189261 & 0.045 \\
Platelets & 205492 & 0.049 \\
Potassium & 241110 & 0.057 \\
Prothrombin time & 139231 & 0.033 \\
Red blood cell count (blood) & 194997 & 0.046 \\
Sodium & 229893 & 0.055 \\
Systolic blood pressure & 3930865 & 0.935 \\
Temperature (C) & 797435 & 0.19 \\
White blood cell count (blood) & 196268 & 0.047 \\
CO2 (ETCO2, PCO2, etc.) & 263161 & 0.063 \\
Oxygen saturation & 101518 & 0.024 \\
Partial pressure of carbon dioxide & 263153 & 0.063 \\
Partial thromboplastin time & 149675 & 0.036 \\
pH (blood) & 285076 & 0.068 \\
Bilirubin (total) & 47707 & 0.011 \\
Lactate & 84510 & 0.02 \\
Lactic acid & 89347 & 0.021 \\
Positive end-expiratory pressure & 53689 & 0.013 \\
Fraction inspired oxygen & 375335 & 0.089 \\
Calcium ionized & 140283 & 0.033 \\
Alanine aminotransferase & 46850 & 0.011 \\
Alkaline phosphate & 45809 & 0.011 \\
Asparate aminotransferase & 46808 & 0.011 \\
\bottomrule
\end{tabular}
\label{table:time_var_features}
\end{table*}

\begin{table*}[t]
  \centering 
  \caption{Relative action frequency of physician policy in 24h mortality forecasting task} 
\begin{tabular}{l|l}
  \toprule
Feature & Relative action frequency \\ \midrule
Anion gap & 0.0021 \\
Bicarbonate & 0.0118 \\
Blood urea nitrogen & 0.0022 \\
Calcium (total) & 0.0120 \\
Chloride (blood) & 0.0022 \\
Creatine kinase & 0.0122 \\
Creatinine (blood) & 0.0059 \\
Diastolic blood pressure & 0.0101 \\
Glascow coma scale total & 0.0046 \\
Glucose (blood) & 0.0125 \\
Heart Rate & 0.0022 \\
Hematocrit & 0.0123 \\
Hemoglobin & 0.0642 \\
Magnesium & 0.0029 \\
Mean blood pressure & 0.0085 \\
Mean corpuscular hemoglobin & 0.0148 \\
Phosphate & 0.0662 \\
Platelets & 0.0143 \\
Potassium & 0.0112 \\
Prothrombin time & 0.0023 \\
Red blood cell count (blood) & 0.0024 \\
Sodium & 0.0122 \\
Systolic blood pressure & 0.0635 \\
Temperature (C) & 0.0111 \\
White blood cell count (blood) & 0.0029 \\
CO2 (ETCO2, PCO2, etc.) & 0.0059 \\
Oxygen saturation & 0.0073 \\
Partial pressure of carbon dioxide & 0.0103 \\
Partial thromboplastin time & 0.0116 \\
pH (blood) & 0.0009 \\
Bilirubin (total) & 0.0134 \\
Lactate & 0.0068 \\
Lactic acid & 0.0112 \\
Positive end-expiratory pressure & 0.0126 \\
Fraction inspired oxygen & 0.0643 \\
Calcium ionized & 0.0133 \\
Alanine aminotransferase & 0.0193 \\
Alkaline phosphate & 0.0112 \\
Asparate aminotransferase & 0.0075 \\
\bottomrule
\end{tabular}
\label{table:time_var_features_24h_mortality}
\end{table*}

\section{Classifier Training Details and Performances}
\label{appendix:classifier_performance}
We train the RNN as follows.
We use LSTM with $1$ hidden layer of $32$ nodes. We regularize the neural network with $\lambda=1$e$-5$ as $\ell_2$ regularization and dropout rate of $0.3$ for input and output layer, and $0.5$ for the hidden layer. 
We use mean imputation (set the missing value as the feature mean) for the time-series features, and add missingness indicators for each feature \citep{lipton2016modeling}.
We discretize the time series into $1$-hour interval and take average value if there are multiple measurements per interval. The RNN takes these $1$-hour discretized grid point for up to $24$ hours time point to classify.
We train two other baselines: Logistic Regression (LR) with $\ell_2$ regularization as $\lambda=1$e$-5$ (selected by cross validation), and Random Forest (RF) with $500$ trees. We concatenate all the features, as long as missingness indicators across all the time points, resulting in $24 * (39 * 2) + 38 = 1910$ features.
We also use mean imputation for these features.


\begin{table}[ht]
  \centering 
  \caption{Hyperameters and ranges for Dueling DQN} 
\begin{tabular}{l|l}
  \toprule
Parameter & Range \\ \midrule
Num. of representation layers & $\{1, 2, 3, 4\}$ \\
Num. of dueling layers & $\{1, 2, 3, 4\}$ \\
Dim. of NN layers & $\{16, 32, 64, 128\}$ \\
Learning rate & $\{5e-2, 1e-3, 5e-3, 1e-4, 5e-4, 1e-5, 5e-5, 1e-6\}$ \\
L2 reg. constant & $\{5e-1, 1e-1, 5e-2, 1e-2, 5e-3, 1e-3, 5e-4, 1e-4\}$ \\
Dropout keep prob. & $\{1.0, 0.9, 0.8, 0.7, 0.6, 0.5\}$ \\
Training batch size & $\{32, 64, 128, 256, 512\}$ \\
Action cost coefficient $\lambda$ & $\{1e-4, 5e-4, 1e-3, 5e-3, 1e-2\}$ \\
\bottomrule
\end{tabular}
\label{table:dqn_hyperparameters}
\end{table}

\begin{table}[ht]
  \centering 
  \caption{Hyperameters and ranges for information gain estimator in OPPE} 
\begin{tabular}{l|l|l}
  \toprule
Parameter & Range & The best model \\ \midrule
Num. of representation layers & $\{1, 2, 3, 4\}$ & 1 \\
Dim. of NN layers & $\{16, 32, 64, 128, 256, 512\}$ & 64 \\
Learning rate & $\{1e-2, 1e-3, 1e-4, 1e-5, 1e-6, 1e-7\}$ & 1e-3 \\
L2 reg. constant & $\{1e-2, 1e-3, 1e-4, 1e-5, 1e-6, 1e-7\}$ & 1e-4 \\
Dropout keep prob. & $\{1.0, 0.9, 0.8, 0.7, 0.6, 0.5\}$ & 0.7 \\
Training batch size. & $\{64, 128, 256, 512, 1024\}$ & 64 \\
\bottomrule
\end{tabular}
\label{table:info_gain_est_hyperparameters}
\end{table}

\end{document}